\def\year{2021}\relax
\documentclass[letterpaper]{article} 
\usepackage{aaai21}  
\usepackage{times}  
\usepackage{helvet} 
\usepackage{courier}  
\usepackage[hyphens]{url}  
\usepackage{graphicx} 
\usepackage{natbib}  
\usepackage{caption} 
\usepackage[switch,mathlines]{lineno}
\usepackage{amsfonts}       
\usepackage{amsmath}
\usepackage{amssymb}
\usepackage{enumitem}
\usepackage{algorithm}
\usepackage{algorithmicx}
\usepackage{algpseudocode}
\usepackage{caption}
\usepackage{subcaption}
\usepackage{bm}
\usepackage{pifont}
\DeclareMathOperator*{\argmax}{arg\,max}

\usepackage{xspace}
\newcommand{\nl}{NL\xspace}
\newcommand{\rl}{RL\xspace}
\newcommand{\LangGoalIRL}{LangGoalIRL\xspace}

\title{Inverse Reinforcement Learning with Natural Language Goals}
\author{Li Zhou, Kevin Small \\}
\affiliations{
Amazon Alexa \\
\{lizhouml, smakevin\}@amazon.com \\
}

\begin{document}
\maketitle

\begin{abstract}
Humans generally use natural language (\nl) to communicate task requirements to each other. Ideally, \nl should also be usable for communicating goals to autonomous machines (e.g., robots) to minimize  friction in task specification. However, understanding and mapping \nl goals to sequences of states and actions is challenging. Specifically, existing work along these lines has encountered difficulty in generalizing learned policies to new \nl goals and environments. In this paper, we propose a novel adversarial inverse reinforcement learning algorithm to learn a language-conditioned policy and reward function. To improve generalization of the learned policy and reward function, we use a variational goal generator to relabel trajectories and sample diverse goals during training. Our algorithm outperforms multiple baselines by a large margin on a vision-based \nl instruction following dataset (Room-2-Room), demonstrating a promising advance in enabling the use of \nl instructions in specifying agent goals.
\end{abstract}

\section{Introduction}
Humans use natural language (\nl) to communicate with each other. Accordingly, a desired method for conveying goals or assigning tasks to an autonomous machine (e.g., a robot) is also by \nl. For example, in household tasks, when asking an agent to pick up a toolbox from the garage, we do not assign the coordinates of the toolbox to the robot. Instead, we state to the agent \textit{retrieve my toolbox from the garage}. This requires the agent to understand the semantics of the natural language goals, associate states and actions with the natural language goals, and infer whether the natural language goals are achieved or not -- all of which are very challenging tasks. Furthermore, we want to be able to learn policies that are robust to lexical variation in goals and generalize well to new goals and environments.

Inverse reinforcement learning (IRL)~\citep{abbeel2004apprenticeship,fu2018learning}, a form of imitation learning~\citep{osa2018algorithmic}, is the task of learning a reward function and hence a policy based on expert demonstrations.
Imitation learning has been successfully applied to a wide range of tasks including robot manipulation~\citep{finn2016guided}, autonomous driving~\citep{kuefler2017imitating}, human behavior forecasting~\citep{rhinehart2017first}, video game AI~\citep{tucker2018inverse}. Generally, task goals are specified intrinsically by the environment and an agent is trained for each task. To generalize the learned policy to new goals, goal-conditioned imitation learning~\citep{ding2019goal} and reinforcement learning (\rl) algorithms~\citep{kaelbling1993learning,schaul2015universal,nair2018visual} have been proposed where the policy is explicitly conditioned on a goal. Normally, the goals either share the same space with the states or can be easily mapped to the state space. For example, in~\citet{ding2019goal}, the goals and states are both coordinates in the environment and the goals are provided to the agent by specifying goal positions.

\citet{fu2019language} and \citet{bahdanau2018learning} represent two recent related works from the perspective of understanding \nl goal specifications, where they propose learning a language-conditioned reward function under the maximum entropy inverse reinforcement learning~\citep{ziebart2008maximum} and generative adversarial imitation learning frameworks~\citep{ho2016generative}. However, the \nl goals are produced by a preset template-based grammar. \citet{fu2019language}'s policy is optimized exactly in a grid environment with known dynamics and when the policy is optimized by sample-based reinforcement learning algorithms such as deep Q-learning~\citep{mnih2015human}, the model performance drops significantly. This reveals a sample efficiency challenge in this setting and implies that when the \nl goals and the environments are complicated, generalization to new goals and environments becomes a very difficult problem.

One potential method for addressing these shortcomings are goal relabeling techniques such as hindsight experience replay (HER)~\citep{andrychowicz2017hindsight} and latent goal relabeling~\citep{nair2018visual}, which have been shown to improve sample efficiency in \rl settings. However, when the goals are \nl and are in a different space than the state space, goal relabeling cannot be applied directly as we cannot easily relabel a state to a \nl goal. \citet{cideron2019self} proposes to build such a relabeling function with a \textsc{Seq2Seq} model that takes a trajectory as input and outputs a relabeled \nl goal. However, this uses the ground-truth reward function and their experiments are again based on simple template-based \nl goals. Moreover, as we will show in this paper, applying HER to IRL with \nl goals doesn't significantly improve performance, as the reward function does not generalize well to relabeled goals. 

In this work, we propose a sample efficient IRL algorithm with natural language (\nl) goals. To the best of our knowledge, our work is the first IRL algorithm that works with real human generated \nl goals (as opposed to template-based language goals~\citep{fu2018learning,bahdanau2018learning}) in a real world vision-based environment. Our contributions include: (1) proposing a \nl goal-conditioned adversarial inverse reinforcement learning algorithm, (2) specifying a variational goal generator that efficiently sample diverse \nl goals given a trajectory, (3) utilizing goal relabeling and sampling strategies to improve the generalization of both the policy and the reward function to new \nl goals and new environments, (4) proposing a self-supervised learning mechanism to further improve the generalization in new environments. Through these innovations, we show that our algorithm significantly outperform strong baselines on the Room-2-Room dataset~\citep{anderson2018vision}.

\section{Problem Formulation}
A task is defined as a pair $(E, G)$, where $E$ is an environment that the agent can interact with and $G$ is a \nl goal that the agent has to fulfill. $G=\{w_1, w_2, ..., w_N\}$ consists of $N$ words. For example, in our experiments, $E$ is a realistic 3D indoor environment and $G$ is a human-generated navigation instruction. The true reward function of each task is unknown, but there exists a set of expert demonstrations consisting of a task ($E$, $G$) and a trajectory $\tau$. The trajectory $\tau = \{s_1, a_1, s_2, a_2, ..., s_T, a_T\}$ consists of a sequence of $T$ states perceived by the experts and the actions taken by the experts given the states. For example, in our experiments the states are first-person views in a 3D indoor environment and the actions are movements towards a direction in the environment. While our proposed methods can be applied to both continuous and discrete action spaces, we focus on discrete action spaces in this paper. The objective of the algorithm is to learn a policy that imitates expert demonstrations such that given a new \nl goal in an existing or new environment, the agent can perform a sequence of actions to achieve that goal. In this paper, we use $G$ to represent an arbitrary goal, $\bm{G}$ to represent the set of goals in expert demonstrations, and $G_i$ to represent the $i$-th goal in $\bm{G}$. The same notation style applies to $E$ and $\tau$ too.

\section{Method}
\subsection{Preliminary: Adversarial Inverse Reinforcement Learning (AIRL)}
AIRL~\cite{fu2018learning} is based on maximum entropy inverse reinforcement learning (MaxEntIRL)~\cite{ziebart2008maximum}. In MaxEntIRL, the probability of a trajectory is defined as 
    $p_\theta(\tau) = \frac{1}{Z} \exp(f_\theta(\tau))$,
where $f_\theta$ is the reward function to learn and $Z = \sum_{\tau} \exp(f_{\theta}(\tau))$ is the partition function. MaxEntIRL learns the reward function from expert demonstrations by maximizing the likelihood of trajectories in the expert demonstrations:
    $\max_\theta \mathbb{E}_{\tau \sim \mathcal{D}}(\log p_\theta(\tau))$.
Maximizing this likelihood is challenging because the partition function $Z$ is hard to estimate. AIRL maximizes this likelihood by using a generative adversarial network (GAN)~\cite{goodfellow2014generative}. The GAN generator is the policy $\pi$ to learn and the GAN discriminator is defined as
\begin{linenomath}
\postdisplaypenalty=0
\begin{align}
    D_{\theta, \phi}(s, a, s') = \frac{\exp \{ f_{\theta, \phi}(s, a, s') \}}{\exp \{ f_{\theta, \phi}(s, a, s') \} +\pi(a|s)}
    \label{eq:discri}
\end{align}
\end{linenomath}
where $f_{\theta, \phi}(s, a, s') = g_\theta(s, a) + \gamma h_\phi(s') - h_{\phi}(s)$, $g_\theta(s, a)$ is the reward function to learn, and $h_{\phi}(s') - h_\phi(s)$ is the reward shaping term. The policy $\pi$ and the discriminator are updated alternately. 

\subsection{Inverse Reinforcement Learning with Natural Language Goals}
\label{sec:ilnlg}
Our AIRL-based solution builds on language-conditioned reward learning (LC-RL)~\citep{fu2019language} by adding template-free \nl capabilities and continuous state space support. In this problem setting, the policy and the reward function are conditioned on a \nl goal $G$, so we first extend the discriminator in Equation (\ref{eq:discri}) to have
$f_{\theta, \phi}(s, a, s', G) = g_\theta(s, a, G) + \gamma h_\phi(s', G) - h_\phi(s, G)$ such that
\begin{linenomath}
\postdisplaypenalty=0
\begin{align*}
    g_\theta(s, a, G) &= \text{MLP}([e^s; \text{Att}(s, G); e^a]) \\
    h_\phi(s, G) &= \text{MLP}([e^s; \text{Att}(s, G)])
\end{align*}
\end{linenomath}
where $\text{MLP}(\cdot)$ is a multilayer Perceptron, $e^s$ and $e^a$ are the embeddings of state $s$ and action $a$, respectively, and $\text{Att}(s, G)$ is an attention function. $\text{Att}(s, G) = \sum \alpha_i e^w_i$ where $e^w_i$ is the word embedding of $w_i$ in $G$, $\alpha_i = \left(\text{Linear}(e^s) \cdot e^w_i \right) / \left(\sum_i \text{Linear}(e^s) \cdot e^w_i \right)$ and $\text{Linear}(\cdot)$ is a single-layer Perceptron.
We use soft actor-critic (SAC)~\cite{haarnoja2018soft}, one of the state-of-the-art off-policy \rl algorithms, to optimize policy $\pi$ given the reward function $g_\theta(s, a, G)$. SAC includes a policy network to predict an action given a state and a goal, and a Q-network that estimate the Q-value of an action given a state and a goal. We define the policy network as
\begin{linenomath}
\postdisplaypenalty=0
\begin{align*}
    \pi_{w}(s, a, G) = \text{Softmax}(e^a \cdot \text{MLP}([e^s; \text{Att}(s, G)]))
\end{align*}
\end{linenomath}
and we define the Q-network $q_{\psi}(s, a, G)$ as the same network architecture as $g_\theta$.
Compared with on-policy algorithms such as TRPO~\cite{schulman2015trust}, SAC utilizes a replay buffer to re-use sampled trajectories. The replay buffer is beneficial to training both the discriminator and the policy. \textit{To update the discriminator}, we sample negative $(s, a, s', G)$ examples from the replay buffer and sample positive $(s, a, s', G)$ examples from the expert demonstrations. \textit{To update the policy},
we sample a batch of $(s, a, s', G)$ from the replay buffer and use $g_\theta$ to estimate their rewards; then we update the Q- and policy network using these reward-augmented samples.
We modify SAC slightly to support discrete action spaces. Appendix \ref{appendix:impdetail} contains details about model architecture and optimization. 

\subsection{A Variational Goal Generator for Improved Generalization and Sample Efficiency}
\label{sec:vgg}
While SAC has better sample efficiency than many on-policy RL algorithms, as we will show in our experiments, in environments with complicated \nl goals and high dimensional state spaces (e.g.,  vision-based instruction following), AIRL with SAC performs only slightly better than supervised learning based behavior cloning and the learned policy and discriminator doesn't generalize well to new goals or new environments. In this section, we show that by learning a variational goal generator, we can enrich the training data for both the discriminator and the policy, which leads to large improvements in sample efficiency and generalization for both the discriminator and policy in \nl settings.

A goal generator takes a trajectory $\tau$ (sequence of states and actions) as input and outputs a \nl goal $G$. Given expert demonstrations, a straightforward way of learning a goal generator is to train an encoder-decoder model that encodes a trajectory and decodes a \nl goal. However, in reality, there are many possible ways to describe a specific trajectory using \nl, and the description will likely be biased due to variance in people's expression preferences. For example, consider a trajectory in which the agent goes to the kitchen and washes the dishes. Two possible goal descriptions for this trajectory can be \textit{go to the kitchen and wash the dishes} or \textit{clean the dishes on the dining table}. To better model variations and generate more diverse \nl goals, we learn a variational encoder-decoder model as goal generator. 
The generative process of the variational goal generator is
\begin{linenomath}
\postdisplaypenalty=0
\begin{gather*}
    \mu_{prior}, \sigma^2_{prior} = f_{prior}\left(\tau \right) \\
    z \sim \mathcal{N}\left(\mu_{prior}, \sigma_{prior}^2\mathbf{I}\right) \\
    G = f_{dec}\left(z, \tau \right)
\end{gather*}
\end{linenomath}
where $f_{prior}$ is a LSTM-based trajectory encoder and $f_{dec}$ is an attention-based goal decoder. The posterior distribution of latent variable $z$ is approximated by
\begin{linenomath}
\postdisplaypenalty=0
\begin{gather*}
    \mu_{posterior}, \sigma_{posterior}^2 = f_{posterior}\left(\tau, G \right)\\
    q\left(z|\tau, G\right) = \mathcal{N}\left(z | \mu_{posterior}, \sigma^2_{posterior}\mathbf{I}\right)
\end{gather*}
\end{linenomath}
where $f_{posterior}$ is a LSTM-based trajectory and goal encoder.
We then maximize the variational lower bound $-D_{KL}(q(z|\tau, G)||p(z)) + \mathbb{E}_{q(z|\tau, G)} [\log p(G|z, \tau)]$.
Network architecture and optimization details are in Appendix \ref{appendix:impdetail}. Given the learned variational goal generator, we propose using three goal relabeling and sampling strategies to improve generalization as itemized below.

\subsubsection{Expert Goal Relabeling (EGR).}
As discussed previously, multiple goals can be mapped to the same trajectory. We propose to augment expert demonstrations by generating $N$ other goals for each expert trajectory $\tau_i$: $G_{i, n} \sim \text{GoalGenerator}(\tau_i),\ n=\{1, 2, ..., N\}$, where $G_{i, n}$ is the $n$-th generated goal for $\tau_i$, and $\text{GoalGenerator}(\cdot)$ is the learned variational goal generator.
For expository and experimental simplicity, we set $N=2$, which works well for this dataset. Model performance under different values of $N$ are shown in Appendix \ref{appendix:addexp}.
These newly generated tuples $\{(E_i, G_{i, n}, \tau_i)\}_{n=1}^N$ are also treated as expert demonstrations for training. We represent the set of goals in the augmented expert demonstrations by $\bm{G}^1$ where the superscript of $\bm{G}$ represents the round of generation as described shortly.

\subsubsection{Hindsight Goal Relabeling (HGR) for the Discriminator.}
\label{sec:hgrd}
In the AIRL framework, the quality of the discriminator is crucial to the generalization of the learned policy. A good discriminator learns a good reward function that generalizes well to new goals and new environments such that the learned policy can also well generalize to new goals and new environments~\cite{fu2019language}. To improve the discriminator, we propose to augment the positive examples of the discriminator by relabeling the goals of the sampled trajectories. More specifically, during the training, given a goal $(E_j, G^1_j)$, the agent interacts with the environment and samples a trajectory $\tau^1_j$. We then use the variational goal generator to sample a goal for $\tau^1_j$:
\begin{linenomath}
\postdisplaypenalty=0
\begin{align}
\tau^1_j &\sim \pi(E_j, G^1_j) \label{eq:g1} \\
G^2_j &\sim \text{GoalGenerator}(\tau^1_j)
\end{align}
\end{linenomath}
The tuples $(\bm{E}, \bm{G}^2, \bm{\tau}^1)$ are treated as positive examples for the discriminator, as a supplement to the positive examples from expert demonstrations $(\bm{E}, \bm{G}^1, \bm{\tau})$. 

\subsubsection{Hindsight Goal Sampling (HGS) for Policy Optimization.}
When optimizing the policy with SAC, we have to sample \nl goals to train the policy. One natural way is to sample goals from expert demonstrations. However, expert demonstrations are relatively scarce, expensive to acquire, and it is difficult to cover goal variance encountered in the testing. Meanwhile, as we will show in our experiments, training with a diverse set of goals is beneficial to the generalization of the policy. Therefore, we propose to also sample goals from $\bm{G}^2$ so that the policy can train with goals beyond these in the expert demonstrations.
\begin{linenomath}
\postdisplaypenalty=0
\begin{align}
\tau_j^2 &\sim \pi(E_j, G^2_j) \\
G^3_j &\sim \text{GoalGenerator}(\tau^2_j)
\label{eq:g3}
\end{align}
\end{linenomath}
Of course, training the policy with $\bm{G}^2$ relies on the discriminator's generalization ability to provide reasonable reward estimates for states and actions sampled under $\bm{G}^2$. This is ensured by HGR in Section \ref{sec:hgrd}, as $(\bm{E}, \bm{G}^2, \bm{\tau}^1)$ are provided as positive examples for the discriminator.

The process from Equation (\ref{eq:g1}) to Equation (\ref{eq:g3}) can be seen as using $\bm{G}^1$ and $\bm{\tau}$ as seed, and iteratively sample $\bm{\tau}^{v}$ from $\bm{G}^v$ using the policy, and then sample $\bm{G}^{v+1}$ from $\bm{\tau}^{v}$ using the goal generator. We can go deeper in this loop to sample more diverse \nl goals and trajectories for policy and discriminator training. More specifically, \textit{to train the discriminator}, positive examples are sampled from $(\bm{E}, \bm{G}^1, \bm{\tau})$ with probability $0.5$, and are sampled from $\{(\bm{E}, \bm{G}^{v+1}, \bm{\tau}^v)|v \ge 1\}$ with probability $0.5$; negative examples are sampled from $\{(\bm{E}, \bm{G}^v, \bm{\tau}^v)|v \ge 1\}$. \textit{To train the policy}, goals are sampled from $(\bm{E}, \bm{G}^1)$ with probability $0.5$ and are sampled from $\{(\bm{E}, \bm{G}^{v+1})| v\ge 1\}$ with probability $0.5$.
Algorithm \ref{algo:algo} shows the overall description for LangGoalIRL.

\subsubsection{Hindsight Experience Replay (HER) for Policy Optimization.}
A closely related approach, hindsight experience replay~\citep{andrychowicz2017hindsight}, has been shown to work very well in \rl settings with sparse rewards. In principle, we could easily incorporate HER into our training procedure. That is, when sampling from replay buffer to optimize policy $\pi$, we sample batches not only from $\{(\bm{E}, \bm{G}^v, \bm{\tau}^{v})\ | v \ge 1\}$, but also from $\{(\bm{E}, \bm{G}^{v+1}, \bm{\tau}^v)\ | v \ge 1\}$. The difference between HER and HGR is that the former is goal relabeling for the policy while the latter is goal relabeling for the discriminator. 
HER is most efficient when rewards are sparse; however, in our setting, the rewards provided by the discriminator are not sparse and we do not observe a boost of performance after applying HER. This is discussed in greater detail in Section~\ref{sec:experiments}.

\algnewcommand{\LineComment}[1]{\State \(\triangleright\) #1}
\renewcommand{\algorithmicrequire}{\textbf{Input:}}
\begin{algorithm}[!t]
\footnotesize
\caption{Inverse Reinforcement Learning with Natural Language Goals (LangGoalIRL)}
\begin{algorithmic}[1]
  \Require GoalGenerator: the variational goal generator from section \ref{sec:vgg}; $\mathcal{D}$: expert demonstrations; $\mathcal{R}=\varnothing$: replay buffer; $b$: batch size; N: number of expert relabeling goals.
  \State $\tilde{\mathcal{D}} = \varnothing$, $\mathcal{G} = \varnothing$
  \For{each $(E_i, G_i, \tau_i) \in \mathcal{D}$} 
  \For{n = 1, 2, ..., N}
  \LineComment{\textbf{Expert Goal Relabeling (EGR)}}
  \State $G_{i, n} \sim \text{GoalGenerator}(\tau_i)$
  \EndFor
  \State add $(E_i, G_i, \tau_i)$ and $\{(E_i, G_{i,n}, \tau_i)\}_{n=1}^N$ to $\tilde{\mathcal{D}}$
  \EndFor
  \While{not converged}
  \State $r_1, r_2 \sim \text{Uniform}(0, 1)$
  \If{$r_1 < 0.5$}
  \State Sample a goal $(E_j, G_j) \sim \tilde{\mathcal{D}}$
  \Else
  \LineComment{\textbf{Hindsight Goal Sampling (HGS)}}
  \State Sample a goal $(E_j, G_j) \sim \mathcal{G}$ 
  \EndIf
  \State Sample a trajectory $\tau'_j \sim \pi(E_j, G_j)$
  \State Sample a relabeled goal $G'_j \sim \text{GoalGenerator}(\tau'_j)$
  \State Add $(E_j, G_j, G'_j, \tau'_j)$ to replay buffer $\mathcal{R}$
  \State Add $(E_j, G'_j)$ to $\mathcal{G}$ 
  \If{$r_2 < 0.5$}
  \State Sample a batch $\mathcal{P}_{+} = \{(s_k^t, a_k^t, s_k^{t+1},G_k)\}_{k=1}^{b} \sim \tilde{\mathcal{D}}$
  \Else
  \LineComment{\textbf{Hindsight Goal Relabeling (HGR)}}
  \State Sample a batch $\mathcal{P}_{+} = \{(s_k^t, a_k^t, s_k^{t+1},G'_k)\}_{k=1}^{b} \sim \mathcal{R}$ 
  \EndIf
  \State Sample a batch $\mathcal{P}_{-} = \{(s_k^t, a_k^t, s_k^{t+1}, G_k)\}_{k=1}^{b} \sim \mathcal{R}$ 
  \State Update discriminator parameters with $\mathcal{P}_{+}$ and $\mathcal{P}_{-}$ as positive and negative examples, respectively.
  \State Sample a batch $\mathcal{Q} = \{(s_k^t, a_k^t, s_k^{t+1}, G_k)\}_{k=1}^{b} \sim \mathcal{R}$
  \State Expand each entry of $\mathcal{Q}$ with reward $r_k^t = g_\theta(s_k^t, a_k^t, G_k)$ 
  \State Optimize $\pi$ using Soft Actor-Critic with $\mathcal{Q}$
  \EndWhile
\end{algorithmic}
\label{algo:algo}
\end{algorithm}

\subsection{Self-Supervised Learning in New Environments}
\label{sec:ss}
In this section, we specifically examine the scenario where the learned policy is deployed to a new environment. For example, after training an embodied agent to perform tasks in a set of buildings, we may deploy this agent to a new building with different floor plans. We assume that we have access to these new environments, but we do not have any expert demonstrations nor any \nl goals in new environments. 
Note that we can not directly apply \nl goals from existing environments to new environments, because goals are tied to the environments. For example, in instruction-following tasks, an example goal is \textit{go downstairs and walk pass the living room}. However, there may be no stairs in a new environment. As a result, we can not just sample goals from expert demonstrations to train in the new environments.

The high-level idea of our method is to first learn a policy $\pi_p$ that is goal-agnostic. That is, the policy selects an action purely based on the state without conditioning on a goal. This model gives us a prior of how the agent will act given a state. We use behavior cloning to train this policy on expert demonstrations. More details are available in Appendix \ref{appendix:impdetail}. We then sample trajectories in the new environments using this policy and sample goals of these trajectories using the goal generator:
\begin{linenomath}
\postdisplaypenalty=0
\begin{align*}
    \tau^{new} &\sim \pi_{p}(E^{new}) \\
    G^{new} &\sim \text{GoalGenerator}(\tau^{new})
\end{align*}
\end{linenomath}
$(\bm{E}^{new}, \bm{G}^{new}, \bm{\tau}^{new})$ are treated as expert demonstrations in the new environments. We then fine-tune the discriminator and the policy in new environments using Algorithm~\ref{algo:algo} (w/o expert goal relabeling). The self-supervised learning signals come from both the generated expert demonstrations for the new environments, and the hindsight goal relabeling and sampling proposed in Section \ref{sec:vgg}. The generated expert demonstrations can be seen as seeds to bootstrap hindsight goal relabeling and sampling.

\section{Experiments}
\label{sec:experiments}
While \LangGoalIRL works on any natural language (\nl) goal based IRL problems, our experiments focus on a challenging vision-based instruction following problem. We evaluate our model on the Room-2-Room (R2R) dataset~\citep{anderson2018vision}, a visually-grounded \nl navigation task in realistic 3D indoor environments.
The dataset contains $7{,}189$ routes sampled from $90$ real world indoor environments. A route is a sequence of viewpoints in the indoor environments with the agent's first-person camera views. Each route is annotated by humans with $3$ navigation instructions. The average length of navigation instructions is $29$ words. The dataset is split into train ($61$ environments and $14{,}025$ instructions), seen validation ($61$ environments same as train set, and $1{,}020$ instructions), unseen validation ($11$ new environments and $2{,}349$ instructions), and test ($18$ new environments and $4{,}173$ instructions). We don't use the test set for evaluation because the ground-truth routes of the test set are not released. Along with the dataset, a simulator is provided to allow the embodied agent to interact with the environments. The observation of the agent is the first-person camera view, which is a panoramic image. The action of the agent is the nearby viewpoints that the agent can move to.
For details about the dataset, please see Appendix \ref{appendix:data}.
The R2R dataset has become a very popular testbed for language grounding in visual context~\citep{tan2019learning,wang2019reinforced,fried2018speaker,zhu2020vision}. 

We evaluate the model performance based on the trajectory success rate. Each navigation instruction in the R2R dataset is labeled with a goal position. Following~\citet{anderson2018vision}, the agent is considered successfully reaching the goal if the navigation error (the distance between the stop position and the goal position) is less than 3 meters. Note that the goal position is not available to our model, as we use navigation instructions as goals.

We compare our model with the following baselines: (1) \textbf{LangGoalIRL\_BASE}, which corresponds to our proposed model in Section \ref{sec:ilnlg} {\em without} the goal relabeling/sampling strategies proposed in Section \ref{sec:vgg}. (2) \textbf{Behavior Cloning}, which is imitation learning as supervised learning. The model shares the same architecture as the policy network of LangGoalIRL and is trained to minimize cross entropy loss with actions in the expert demonstrations as labels. (3) \textbf{LC-RL (Sampling)}~\cite{fu2019language}, which also uses MaxEntIRL~\cite{ziebart2008maximum} to learn a reward function. It optimizes the policy exactly in a grid environment, which is not scalable to our experimental setting, so we use AIRL with SAC for LC-RL to optimize the policy. In this case, LC-RL is very similar to LangGoalIRL\_BASE, except that LC-RL simply concatenates state and goal embeddings as input to the policy and reward function, while LangGoalIRL\_BASE uses the attention mechanism $\text{Att}(s, G)$ in Section \ref{sec:ilnlg}. (4) \textbf{AGILE}~\cite{bahdanau2018learning}, which proposes to learn a discriminator (reward function) that predicts whether a state is the goal state for a \nl goal or not. The positive examples for the discriminator are the (\nl goal, goal state) pairs in expert demonstrations, and the negative examples are sampled from the replay buffer of SAC. (5) \textbf{HER}~\cite{andrychowicz2017hindsight,cideron2019self}, which uses our learned variational goal generator to relabel trajectories. This corresponds to the \texttt{final} strategy in~\citet{andrychowicz2017hindsight} applied to LangGoalIRL\_BASE. (6) \textbf{Upper Bound}~\citep{tan2019learning}. Recently, there are a few vision-language navigation models~\cite{tan2019learning,wang2019reinforced,fried2018speaker,zhu2020vision} developed and evaluated on R2R dataset. However, they assume the goal positions or the optimal actions at any states are known, and assume the environments can be exhaustively searched without sample efficiency considerations in order to do training data augmentation. As a result, their problem settings are not really comparable, but we include~\citet{tan2019learning}'s result a potential upper bound. Note that without the additional information described above, their model performance reverts to our \textbf{Behavior Cloning} baseline. We will discuss this more in Section~\ref{sec:related}.
For implementation details of our algorithms and the baselines, please refer to Appendix \ref{appendix:impdetail}.

\begin{figure*}[t]
\captionsetup{font=footnotesize}
\captionsetup[subfigure]{font=scriptsize}
\centering
     \makebox[\linewidth][c]{
     \begin{subfigure}[b]{0.35\textwidth}
         \centering
         \includegraphics[width=\textwidth]{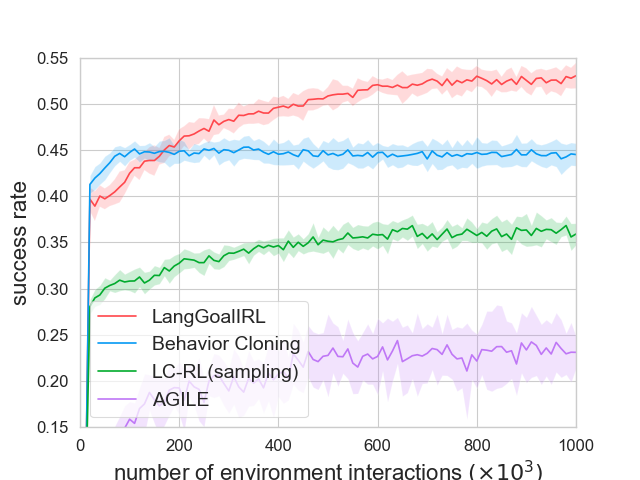}
         \caption{Baseline performance in seen validation.}
     \end{subfigure}
     \begin{subfigure}[b]{0.35\textwidth}
         \centering
         \includegraphics[width=\textwidth]{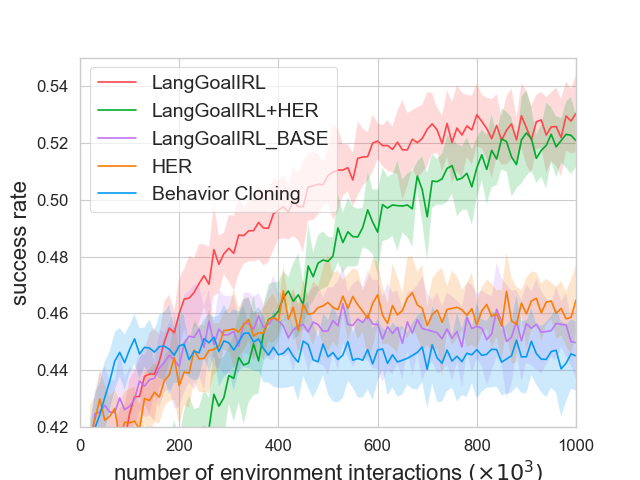}
         \caption{HER variants performance in seen validation.}
     \end{subfigure}
     \begin{subfigure}[b]{0.35\textwidth}
         \centering
         \includegraphics[width=\textwidth]{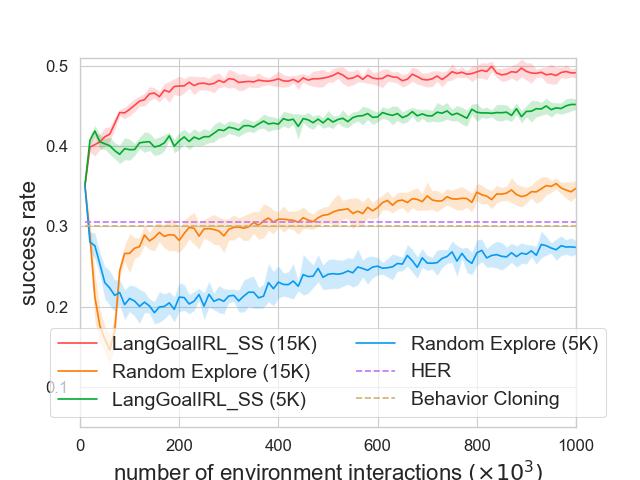}
         \caption{Self-supervised learning in unseen validation.}
     \end{subfigure}
     }
     \makebox[\linewidth][c]{
     \begin{subfigure}[b]{0.35\textwidth}
         \centering
         \includegraphics[width=\textwidth]{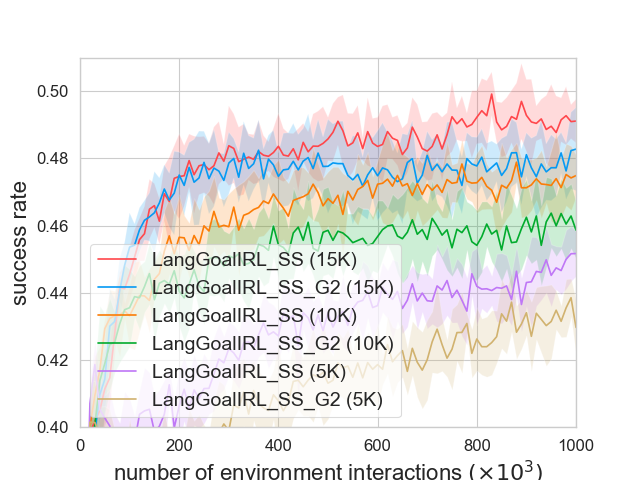}
         \caption{More self-supervised learning in unseen validation.}
     \end{subfigure}
     \begin{subfigure}[b]{0.35\textwidth}
         \centering
         \includegraphics[width=\textwidth]{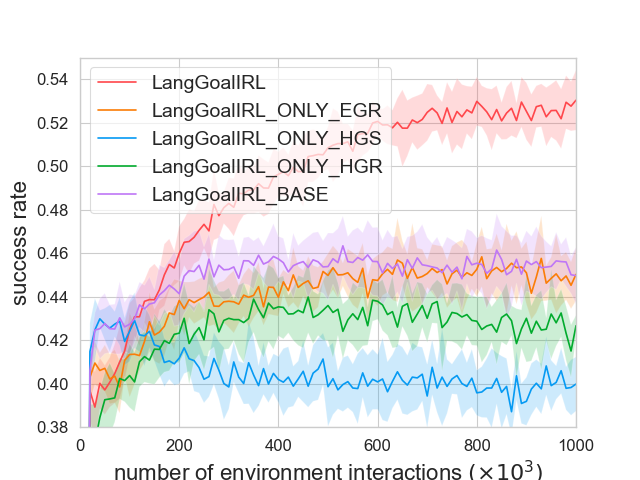}
         \caption{Ablation study in seen validation.}
     \end{subfigure}
     \begin{subfigure}[b]{0.35\textwidth}
         \centering
         \includegraphics[width=\textwidth]{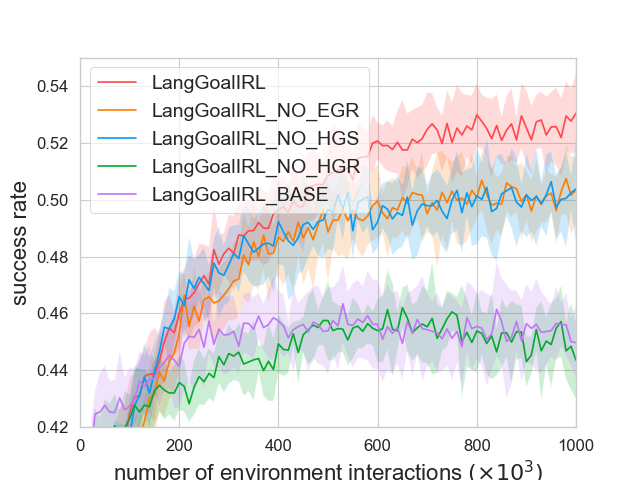}
         \caption{More ablation study in seen validation.}
     \end{subfigure}
     }
     \captionsetup{width=1.0\textwidth}
    \caption{Model performance on both seen and unseen validation. LangGoalIRL\_SS in Figure (c) and (d) represents self-supervised learning in unseen environments. LangGoalIRL\_SS (15K) means that $15,000$ demonstrations are sampled by the goal-agnostic policy $\pi_p$. Random Explore means that a random policy instead of $\pi_p$ is used to sample demonstrations. LangGoalIRL\_SS\_G2 means that the policy does not iteratively sample goals from $\{(\bm{E}, \bm{G}^v) | v > 2\}$ as described in Section 3.3.3.}
    \label{fig:res}
\end{figure*}

The performance of our algorithm and baselines are shown in Figure \ref{fig:res}, Table \ref{tb:res}, and Table \ref{tb:ablation}. From the results, we would like to highlight the following observations.

1.\ \textbf{LangGoalIRL outperforms baselines by a large margin.} From Table \ref{tb:res} and Figure \ref{fig:res}(a) we can see that LangGoalIRL achieves a $0.530$ success rate on seen validation set, a $47.63\%$ improvement over LC-RL (sampling) and a $129\%$ improvement over AGILE. Rewards learned by AGILE are binary and sparse, which may be one of the main reasons why AGILE performs even worse than Behavior Cloning. The difference between LC-RL (sampling) and LangGoalIRL\_BASE is just the lack of attention mechanism over \nl goals, which simply indicates that attention mechanism is important when \nl goals are complicated. Other powerful attention mechanisms such as the Transformer architecture~\cite{vaswani2017attention} could also be easily considered. The success rate of LangGoalIRL is $18.04\%$ and $15.91\%$ higher than LangGoalIRL\_BASE on seen validation and unseen validation, respectively, which shows that the three proposed strategies efficiently improve the generalization of the policy to new \nl goals and new environments. From Figure \ref{fig:res}(b) we see that, when combined with HER, LangGoalIRL converges slower and performs worse than LangGoalIRL alone. HER has been shown to efficiently deal with reward sparsity; however, the rewards learned by AIRL are not sparse. From Table \ref{tb:res} we see that HER alone barely outperforms LangGoalIRL\_BASE. As we will discuss shortly, HER only improves the sample efficiency of the \emph{generator (policy)}, however, the generalization of the \emph{discriminator (reward function)} is what is crucial to the overall performance of the policy. Finally, from Figure \ref{fig:res}(b) we can also see that LangGoalIRL\_BASE consistently outperforms Behavior Cloning after about 200k interactions, which effectively indicates the benefit of using inverse reinforcement learning over behavior cloning.

\begin{table}[t]
\centering
\resizebox{\linewidth}{!}{
\begin{tabular}{l|cc}
\hline
                  & seen validation                 & unseen validation    \\ \hline
Behavior Cloning  & $0.445 \pm 0.0122$              &$0.300 \pm 0.0114$                   \\
LC-RL (sampling)             & $0.359 \pm 0.0117$              & $0.216 \pm 0.0081$                     \\
AGILE             & $0.231 \pm 0.0173$              &$0.253 \pm 0.0269$                      \\
HER               & $0.464 \pm 0.0126$              &$0.306 \pm 0.0173$                      \\ \hline
LangGoalIRL\_BASE & $0.449 \pm 0.0130$              &   $0.308 \pm 0.0087$                   \\
LangGoalIRL       & $\mathbf{0.530 \pm 0.0138}$              &    $0.357 \pm 0.0089$                  \\
\ \ +self-supervised  & \textendash & $\mathbf{0.491 \pm 0.0065}$  \\ \hline
Upper Bound & 0.621 & 0.645 \\ \hline
\end{tabular}
}
\caption{Success rate after 1 million environment interactions. LangGoalIRL+self-supervised is further fine-tuned in unseen environment for 1 million environment interactions (explained in Figure 1(c)).}
\label{tb:res}
\end{table}
\begin{table}[t]
\centering
\resizebox{\linewidth}{!}{
\begin{tabular}{ccc|c}
\hline
\begin{tabular}[c]{@{}c@{}}Expert\\ Goal Relabeling\end{tabular} & \begin{tabular}[c]{@{}c@{}}Hindsight\\ Goal Relabeling\end{tabular} & \begin{tabular}[c]{@{}c@{}}Hindsight\\ Goal Sampling\end{tabular} & \begin{tabular}[c]{@{}c@{}}seen validation\\ success rate\end{tabular} \\ \hline
\textendash                                                      & \textendash                                                         & \textendash                                                       & $0.449 \pm 0.0130$                                                     \\ \hline
\ding{51}                                                        & \textendash                                                         & \textendash                                                       & $0.450 \pm 0.0122$                                                     \\
\textendash                                                      & \ding{51}                                                           & \textendash                                                       & $0.427 \pm 0.0048$                                                     \\
\textendash                                                      & \textendash                                                         & \ding{51}                                                         & $0.399 \pm 0.0124$                                                     \\ \hline
\ding{51}                                                        & \ding{51}                                                           & \textendash                                                       & $0.504 \pm 0.0185$                                                     \\
\ding{51}                                                        & \textendash                                                         & \ding{51}                                                         & $0.444 \pm 0.0134$                                                     \\
\textendash                                                      & \ding{51}                                                           & \ding{51}                                                         & $0.503 \pm 0.0117$                                                     \\ \hline
\ding{51}                                                        & \ding{51}                                                           & \ding{51}                                                         & $\mathbf{0.530 \pm 0.0138}$                                            \\ \hline
\end{tabular}
}
\caption{Ablation study of the three proposed strategies on seen validation set. A complete ablation study which includes HER is in Table \ref{tb:fullablation} of Appendix \ref{appendix:addexp}.}
\label{tb:ablation}
\end{table}

\begin{table*}[]
\resizebox{1.0\linewidth}{!}{
\begin{tabular}{|l|c|c|c|c|c|c|}
\hline
    & LangGoalIRL      & \begin{tabular}[c]{@{}c@{}}Language Goal-Conditioned IRL\\ \citep{fu2019language}\\ \citep{bahdanau2018learning}\end{tabular} & \begin{tabular}[c]{@{}c@{}}Language Goal-Conditioned RL\\ \citep{cideron2019self}\\ \citep{jiang2019language}\end{tabular} & \begin{tabular}[c]{@{}c@{}}State Goal-Conditioned IRL\\ \citep{ding2019goal}\end{tabular} & \begin{tabular}[c]{@{}c@{}}State Goal-Conditioned RL\\ \citep{andrychowicz2017hindsight}\\ \citep{plappert2018multi}\end{tabular} & \begin{tabular}[c]{@{}c@{}}SoTA models on R2R dataset\\ \citep{tan2019learning}\\ \citep{fried2018speaker}\\ \citep{zhu2020vision}\end{tabular} \\ \hline
\begin{tabular}[c]{@{}l@{}}Require Ground-Truth\\ Reward function\end{tabular} & No               & No                                                                                                                                                         & Yes                                                                                                                                                     & No                                                                                                   & Yes                                                                                                                                                           & No (Yes if use RL fine-tuning)                                                                                                                                                                     \\ \hline
Goal Type                                                                      & Natural Language & Natural Language                                                                                                                                           & Natural Language                                                                                                                                        & States                                                                                               & States                                                                                                                                                        & Natural Language                                                                                                                                                                                   \\ \hline
Use Student-forcing                                                            & No               & No                                                                                                                                                         & No                                                                                                                                                      & No                                                                                                   & No                                                                                                                                                            & Yes                                                                                                                                                                                                \\ \hline
\begin{tabular}[c]{@{}l@{}}Goals/Paths\\ Augmentation\end{tabular}             & EGR, HGS, HGR    & No                                                                                                                                                         & HER                                                                                                                                                     & HER                                                                                                  & HER                                                                                                                                                           & \begin{tabular}[c]{@{}c@{}}All shortest-paths between any \\ two locations in environments\end{tabular}                                                                                            \\ \hline
\end{tabular}
}
\caption{We compare our algorithm with closely related works that fall into $5$ different problem settings. In the setting we are interested in, the goals of the tasks are NL goals, and neither the ground-truth reward function (required by RL) nor the optimal action at any state (required by student-forcing) is available.}
\label{tb:assumption_comparison}
\end{table*}
2.\ \textbf{Hindsight goal relabeling and expert goal relabeling improve the generalization of the \emph{discriminator (reward function)}; such generalization is \emph{crucial} to the performance of the policy.} From Table \ref{tb:ablation} and Figure \ref{fig:res}(f), we see that if we take out HGR, the success rate drops significantly from $0.530$ to $0.444$. This shows that HGR plays a key role in learning a better reward function by enriching positive examples of the discriminator. Meanwhile, when applying HGS without HGR, the success rate drops from $0.503$ to $0.399$. This is because the discriminator cannot generalize well to sampled goals when HGR is not applied, which shows the importance of discriminator generalization.
However, from Table \ref{tb:ablation} and Figure \ref{fig:res}(e) we can also see that if we only keep HGR, the success rate drops to $0.427$ which is even lower than LangGoalIRL\_BASE. We observe that in this case the goals generated by the variational goal generator only appear in the positive examples of the discriminator, so the discriminator easily identify these goals, and simply assigns high rewards to these goals regardless of what action is taken. When EGR and HGR are applied together, relabeled goals appear in both positive and negative examples, and the success rate improves from $0.427$ to $0.504$.

3.\ \textbf{Self-supervised learning largely improves performance in unseen environments.} Table \ref{tb:res} shows that with self-supervised learning, LangGoalIRL achieves a success rate of $0.491$, a $59.42\%$ improvement over LangGoalIRL\_BASE and a $37.54\%$ improvement over LangGoalIRL w/o self-supervised learning. This shows that the proposed self-supervised learning algorithm can significantly improve policy performance even though neither expert demonstrations nor \nl goals in new environments are given. The number of trajectories sampled by the goal-agnostic policy $\pi_p$ in new environments by default is $15{,}000$. In Figure \ref{fig:res}(c), we show the performance of a baseline model where the embodied agent randomly explores the new environments (randomly selects an action at any states, but does not visit any visited viewpoints) rather than using $\pi_p$ to sample trajectories. We set the number of sampled trajectory to $15{,}000$ and $5{,}000$. We see that the agent performance drops during early stages, as randomly explored trajectories are not good seeds to bootstrap Algorithm \ref{algo:algo} in new environments. After early stages, the model performance gradually increases as HGR and HGS provide some self-supervised signals but the overall the performance remains lower than \LangGoalIRL. More results are in Table \ref{tb:ss3} of Appendix \ref{appendix:addexp}. Finally, self-supervised learning can also be applied to existing environments by augmenting the expert demonstrations in seen validation set. While this is not our target setting, we include these results in Table \ref{tb:sstrain} of Appendix \ref{appendix:addexp}.

4.\ \textbf{Hindsight goal sampling improves the generalization of the policy by enabling the policy to explore a more diverse set of goals.} From Table \ref{tb:ablation} and Figure \ref{fig:res}(f) we can see that HGS further improves the policy performance from $0.504$ to $0.530$ on seen validation set.
Figure \ref{fig:res}(d) shows the impact of HGS on self-supervised learning in new environments. The baseline, LangGoalIRL\_SS\_G2 samples goals only from $\bm{G}^1$ and $\bm{G}^2$ defined in Section \ref{sec:vgg} and does not iteratively sample goals from $\{(\bm{E}, \bm{G}^v)| v > 2\}$. As goals sampled are less diverse, we see that LangGoalIRL\_SS\_G2 performs worse than LangGoalIRL\_SS given $5{,}000$, $10{,}000$ and $15{,}000$ generated expert demonstrations in new environments.
More experiments are in Table \ref{tb:ss3} of Appendix \ref{appendix:addexp}.

\section{Related Works}
\label{sec:related}
Our paper focuses on proposing a sample efficient algorithm for IRL with natural language (\nl) goals.
Table \ref{tb:assumption_comparison} summarizes the differences between our LangGoalIRL algorithm and recent closely related works on goal-conditioned \rl and imitation learning (IL).
Goal-conditioned \rl and IL have been explored by many prior works~\cite{schaul2015universal,florensa2017automatic,nair2018visual,plappert2018multi,ding2019goal}. Usually, the task is to learn a policy that is parameterized by a goal and can generalize to new goals. 
In most prior works, the goals and states are in a same space, such as positions in a Cartesian coordinate system~\cite{plappert2018multi,ding2019goal}. This corresponds to column $5$ and $6$ in Table \ref{tb:assumption_comparison}.
However, in this paper we investigate a different setting where the goals are \nl goals, which corresponds to column $3$ and $4$ in Table \ref{tb:assumption_comparison}.
The closest prior works to our algorithm are language-conditioned IRL and reward learning~\cite{fu2019language,bahdanau2018learning,MacGlashan-RSS-15,williams2018learning,goyal2019natural}.
\citet{goyal2019natural} propose to train a model to predict whether a language instruction describes an action in a trajectory, and use the predictions to perform reward shaping. \citet{bahdanau2018learning} propose to use GAN~\citep{goodfellow2014generative,ho2016generative} to learn a discriminator that predicts whether a state is the goal state of a \nl instruction. \citet{fu2019language} propose to use the MaxEntIRL~\citep{ziebart2008maximum} framework to learn a language-conditioned reward function for instruction following tasks. The language instructions in \citet{bahdanau2018learning} and \citet{fu2019language}'s experiments are generated by templates and much easier to understand compared with our dataset.
Meanwhile, \LangGoalIRL demonstrates significantly better generalization than these two prior works. There are also many prior works on goal relabeling to improve sample efficiency, but none of them can be directly applied to the IRL with \nl goals setting. \citet{andrychowicz2017hindsight} propose hindsight experience replay that samples additional goals for each transition in the replay buffer and can efficiently deal with the reward sparsity problem. \citet{nair2018visual} propose to sample additional diverse goals from a learned latent space. \citet{ding2019goal} propose to relabel trajectories using the states within the trajectories. These algorithms do not work with \nl goals. \citet{cideron2019self} and \citet{jiang2019language} generalize HER to language setting and relabel trajectories by hindsight language instructions. 
However, in this paper we show that, when doing IRL, simply applying HER to policy optimization does not improve policy performance, and it is critical to improve the generalization of the reward function. 
Finally, there are multiple recent works~\citep{tan2019learning,wang2019reinforced,fried2018speaker,zhu2020vision} on R2R dataset that focus on solving vision-language navigation~\citep{anderson2018vision}. This corresponds to the last column in Table \ref{tb:assumption_comparison}. These works do not focus on sample efficient IRL algorithms and use student-forcing~\citep{anderson2018vision} to train their supervised learning models or pre-train their RL models. With student-forcing, the policy is assumed to always have access to the optimal action at any state it encounters during training, which is even a stronger assumption than knowing the ground-truth reward function.
\citet{fried2018speaker} and \citet{tan2019learning} propose to learn a speaker model that back-translate a given trajectory to a \nl goal, which is similar to our goal generator without latent variables. However, the purpose of the speaker model is to augment training data before the training, rather than relabeling goals during training. To augment training data, they generate the shortest path between any two viewpoints. This requires knowing the topological graph of viewpoints before the training. Moreover, they augment training set with every shortest path that is not included in the original training set. This assumes the environments can be exhaustively searched without sampling efficiency considerations. Other proposed techniques in these papers such as environment dropout~\citep{tan2019learning} and progress monitoring~\citep{zhu2020vision} are orthogonal to our work.
For a survey of reinforcement learning with \nl, please see \citet{luketina2019survey}.

\section{Conclusion}
In this work, we propose a sample efficient algorithm for IRL with \nl goals. Observing limitations of existing works regarding language-conditioned IRL, \LangGoalIRL emphasizes template-free \nl goal specification and sample efficiency when generalizing to new goals and environments. Specifically, we propose learning a variational goal generator that can relabel trajectories and sample diverse goals. Based on this variational goal generator, we describe three strategies to improve the generalization and sample efficiency of the language-conditioned policy and reward function. Empirical results demonstrate that \LangGoalIRL outperforms existing baselines by a large margin and generalizes well to new natural language goals and new environments, thus increasing flexibility of expression and domain transfer in providing instructions to autonomous agents.

\bibliography{arxiv}

\appendix
\section*{Appendix}
\section{Dataset and Environment Details}
\label{appendix:data}
The Room-2-Room (R2R) dataset~\footnote{https://bringmeaspoon.org/}~\citep{anderson2018vision} is a dataset for visually-grounded natural language navigation task in realistic 3D indoor environments. The dataset is built on top of the Matterport3D simulator~\citep{anderson2018vision}. Matterport3D simulator provides APIs to interact with $90$ $3$D indoor environments, including homes, offices, churches and hotels. An agent can navigation between viewpoints in the $3$D environment. At each viewpoint, the agent can turn around and perceive a $360$-degree panoramic image. Images are all real rather than synthetic~\citep{Matterport3D}.

\begin{figure*}[h]
\centering
\begin{subfigure}{0.405\textwidth}
  \centering
  \includegraphics[width=0.95\linewidth]{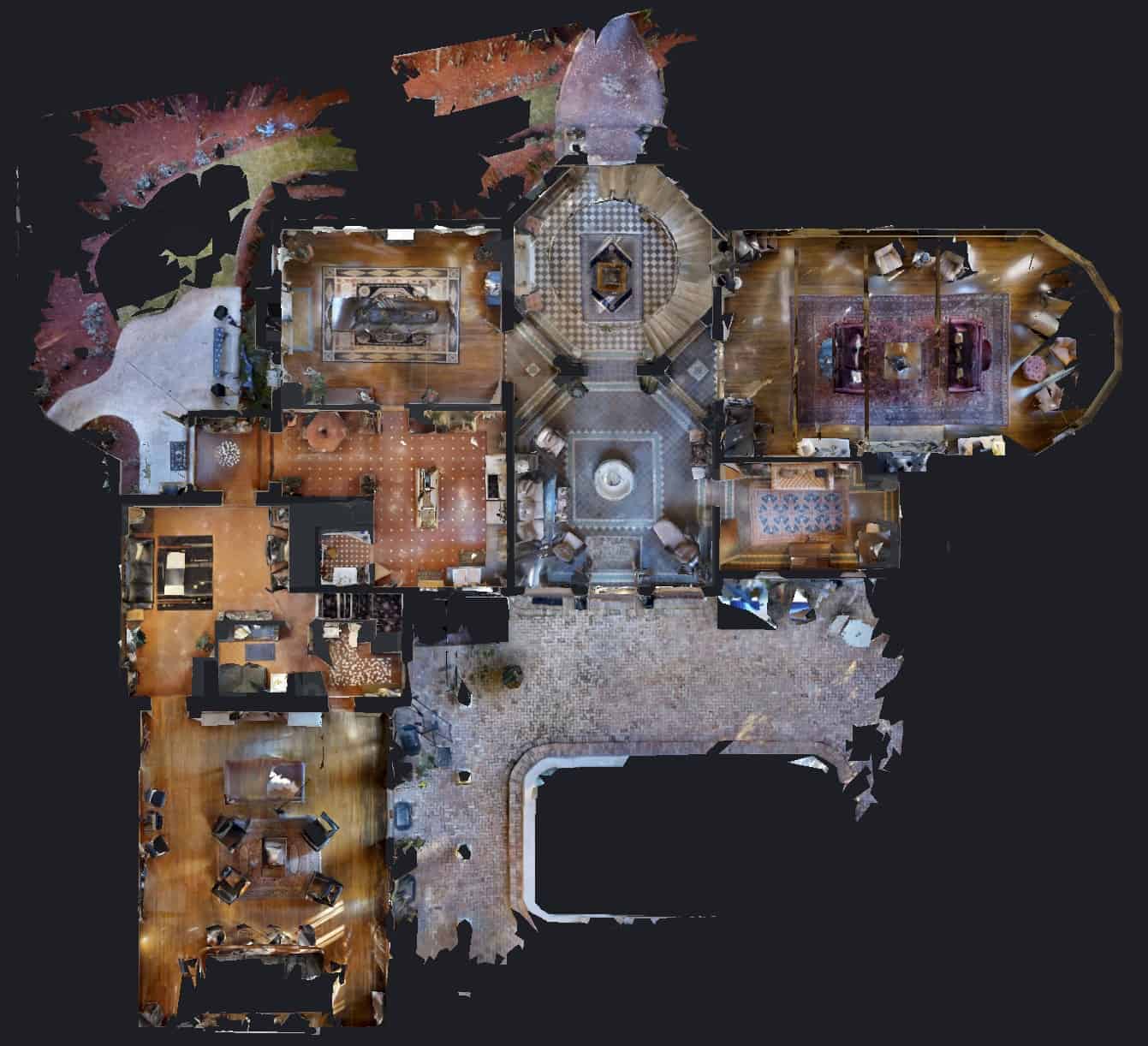}
  \label{fig:birdview}
\end{subfigure}%
\begin{subfigure}{.585\textwidth}
  \centering
  \includegraphics[width=0.95\linewidth]{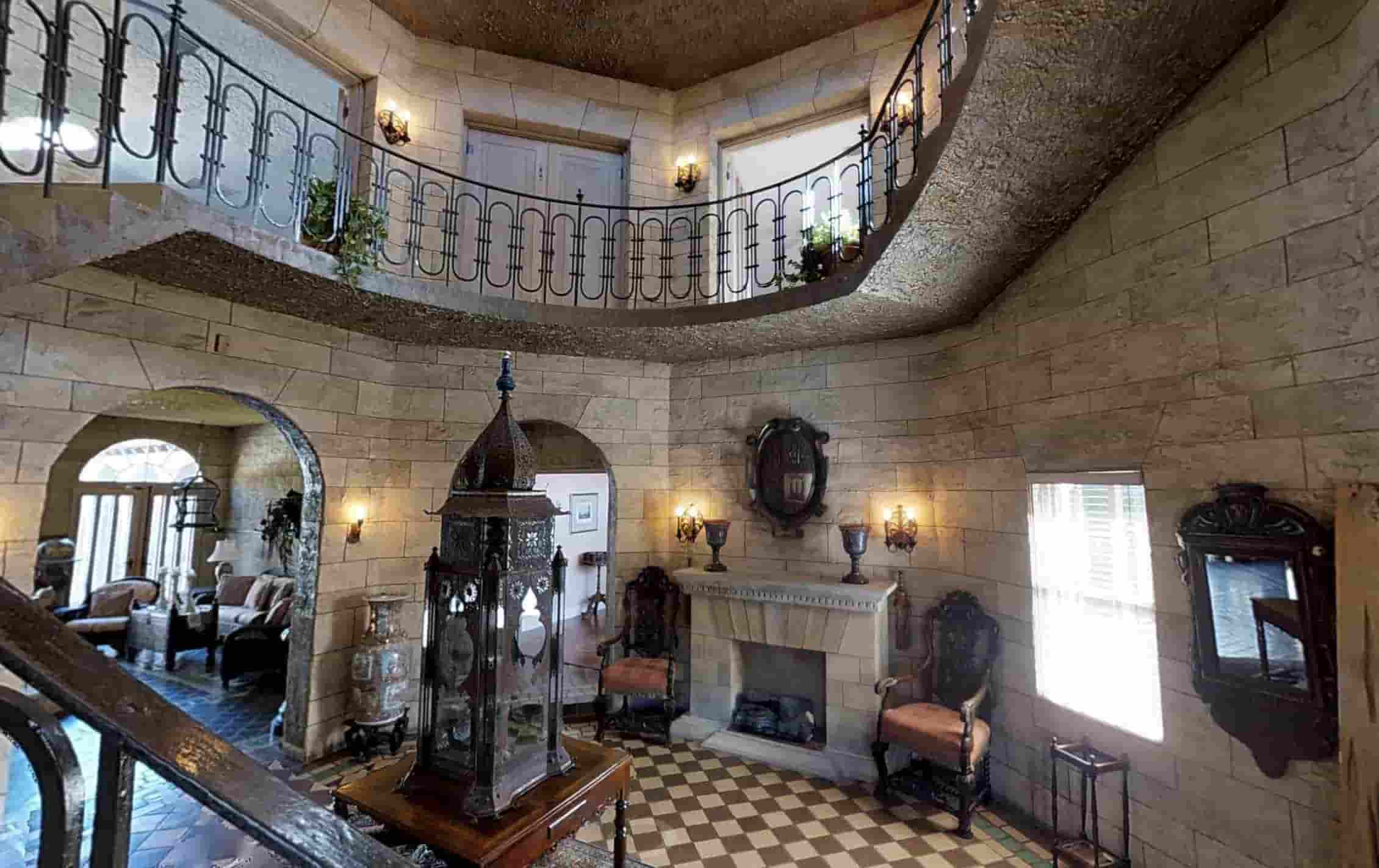}
  \label{fig:cameraview}
\end{subfigure}
\caption{The left-side figure shows a bird view of the indoor environment. This view is for demonstration purpose and is not available to the embodied agent. The right-side figure is the camera view of the embodied agent at a viewpoint. The agent can look around or move to the next viewpoint.}
\label{fig:r2r}
\end{figure*}
The routes in the R2R dataset was sampled from the simulator by first sampling two viewpoints in the environment and then calculating the shortest-path from one viewpoint to the other. In total, there are $7{,}189$ visually diverse routes sampled. Then each routes was annotated by $3$ Amazon Mechanical Turk workers. The workers are asked to write directions to provide navigation instructions for sampled routes so that a robot can follow the instructions to reach to the end viewpoints. In total there are $21{,}567$ instructions, and the average length of each instruction is $29$ words. The size of the vocabulary for training is $991$. The dataset is split into train ($61$ environments and $14{,}025$ instructions), seen validation ($61$ environments same as train set, and $1{,}020$ instructions), unseen validation ($11$ new environments and $2{,}349$ instructions), and test ($18$ new environments and $4{,}173$ instructions). 

The following are $3$ navigation instructions in the dataset describing the same route:
\begin{enumerate}
\item \textit{Head indoors and take the hallway to the living room. Stop and wait on the right hand side of the purple wall.} 
\item \textit{Turn slightly left and walk across the hallway. Turn slightly right and walk towards the green sofas. Walk to the right entrance where the window panes are at and wait there.} 
\item \textit{Go inside and walk straight down the hallway.  Keep going until you see a green couch on the right and then walk towards it.  Go to the right side of the TV and stop in the doorway leading to an interior room with a table in it.}
\end{enumerate}
For more details about the dataset, please see~\citet{anderson2018vision}.
\section{Implementation Details}
\label{appendix:impdetail}
\textbf{Observation Space}: The Matterport3D simulator outputs a RGB image corresponding to the embodied agent's first-person camera view. The image resolution is $640 \times 480$. Following~\citet{fried2018speaker} and~\citet{tan2019learning}, at the each viewpoint, the agent first look around and get a $360$ degree panoramic image, which consists of $36$ first-person camera view images ($12$ heading and $3$ elevation with $30$ degree increments).
Image features are $2048$-dimensional vecotors extracted from a pre-trained ResNet-152 model~\footnote{https://github.com/peteanderson80/Matterport3DSimulator}~\citep{he2016deep}.

\noindent \textbf{State Space}: The observation of the embodied agent only captures views from the current viewpoint. We define the state of the embodied agent as the set of all previous observations in the current trajectory. We use a LSTM model to encode previous and current observations into a $n$-dimensional vector. If we update the LSTM model parameters during training, it would make our problem a POMDP problem~\citep{lovejoy1991survey}. For the purpose of simplicity, we pre-train the LSTM model using the base model below and fix its parameters during training.

\noindent \textbf{Action Space}: The action space at any given state is the collection of nearby viewpoints. The maximum number of nearby viewpoints is $13$. The feature vector of each action (viewpoint) is a $m$-dimensional vector that will be described in the base model below. We add a $m$-dimensional vector of all zeros to represent the stop action, so the maximum number of actions given a state is $14$.

\noindent \textbf{Base Model}: we use a base model architecture that is similar to many prior works~\citep{fried2018speaker,tan2019learning,wang2019reinforced}. The base model is a sequence-to-sequence model. The encoder takes natural language goals as input and the decoder outputs a sequence of actions. Let $\{o_{t, i}\}_{i=1}^{36}$ be the observations ($36$ first-person camera view images) at time step $t$. Let $\theta_{t, i}$ and $\phi_{t, i}$ be the heading and elevation of the camera view $o_{t, i}$. The feature vector of $o_{t, i}$ is $f_{t, i} = [f_{\text{ResNet}}(o_{t, i}); \cos{\theta_{t, i}}; \sin{\theta_{t, i}}; \cos{\phi_{t, i}}; \sin{\phi_{t, i}}]$, where $f_{\text{ResNet}}$ is a pre-trained ResNet-152 model on ImageNet dataset~\footnote{http://image-net.org/index}.
Let the natural language goal be $G=\{w_1, w_2, ..., w_N\}$. The natural language goal is encoded by a LSTM model:
\begin{linenomath}
\postdisplaypenalty=0
\begin{align*}
    h^w_1, h^w_2, ..., h^w_N = \text{LSTM}_{enc}(e^w_1, e^w_2, ..., e^w_N)
\end{align*}
where $e^w_n$ is the $n$-th word embedding. We use GloVe~\cite{pennington2014glove} to initialize the weights of word embeddings. The decoder is also a LSTM model. Let the hidden state of the decoder at step $t$ be $h^s_t$. We use $h^s_{t-1}$ to attend over the observations at $t$
\begin{align*}
\alpha^s_i &= \text{Softmax}(\text{Linear}(h^s_{t-1}) \cdot f_{t, i}) \\
f_t &= \sum_{i=1}^{36} \alpha^s_i f_{t, i}
\end{align*}
Then we update the decoder model:
\begin{align*}
    e^{angle}_t &= \text{Linear}([\cos{\theta_{t}}; \sin{\theta_{t}}; \cos{\phi_{t}}; \sin{\phi_{t}}]) \\
    h^s_{t} &= \text{LSTM}_{dec}(h^s_{t-1}, [f_t; e^{angle}_t])
\end{align*}
where $\theta_t$ and $\phi_t$ are the current heading and elevation of the embodied agent.
To predict the next action, we first use $h^s_t$ to attend over the natural language goal: $\alpha^w_n = \text{Softmax}(\text{Linear}(h^s_{t}) \cdot h^w_{n})$. Then the summarized goal embedding is $g_t =\sum_{n=1}^{N} \alpha^w_n h^w_n$.
Then the probability of selecting an action is given by
\begin{align}
    p^a_i = \text{Softmax}(\text{BiLinear}(\text{Linear}([h_t^s; g_t]), e^a_{t,i}))
    \label{eq:basemodelp}
\end{align}
\end{linenomath}
where $\text{BiLinear}(v, u) = v^\top W u$, $e^a_{t,i} = [f_{\text{ResNet}}(o_{t, a}); \cos{\theta_{t, a}}; \sin{\theta_{t, a}}; \cos{\phi_{t, a}}; \sin{\phi_{t, a}}]$ is the action embedding, $o_{t, a}$ is the camera view of the action (viewpoint) from the current viewpoint, and $\theta$ and $\phi$ are heading and elevation of the action from the current viewpoint. The base model is trained with cross-entropy loss $L_{base} = \text{CrossEntropy}(p^a, y^a)$ where $y^a$ is the ground-truth action. The trained $\text{LSTM}_{dec}$ is used as the pre-trained state encoding model for all algorithms in our experiments.

\noindent \textbf{Policy Model}: We use soft actor-critic (SAC)~\cite{haarnoja2018soft} as our policy optimization algorithm. SAC includes a Q-network and a policy network. The Q  network $Q(s, a, G)$ is defined as
\begin{linenomath}
\postdisplaypenalty=0
\begin{align*}
    Q_{\psi}(s_t, a_{t,i}, G) = \text{MLP}([h_t^s; g_t; e^a_{t,i}])
\end{align*}
where $\text{MLP}$ is a multilayer feedforward neural network, $h_t^s$ is the state of the embodied agent, which is given by the pre-trained $\text{LSTM}_{dec}$ model in the base model, $g_t$ is the summarized goal embedding calculated in the same way as the base model, and $e^a_{t,i}$ is the action embedding. The policy network is defined as 
\begin{align*}
    \pi_w(s_t, a_{t,i}, G) = \text{Softmax}(e^a_{t,i} \cdot \text{MLP}([h_t^s; g_t]))
\end{align*}
By default the action space in SAC is continuous, however, in our setting the action space is discrete and state-conditioned. We modify the SAC algorithm slightly so that it can support discrete action space. The Q-network parameters in SAC are trained to minimize the following soft Bellman residual:
\begin{align*}
    &J_Q(\psi) = \mathbb{E}_{(s_t, a_{t,i}, G)\sim \mathcal{D}} [ \\
    &\quad \frac{1}{2}\left(Q_\psi(s_t, a_{t,i}, G) - \left(r_t + \gamma \mathbb{E}_{s_{t+1}}\left[V_{\bar{\psi}}(s_{t+1}, G)\right]\right)\right)^2] \\
    &]
\end{align*}
where $\bar{\psi}$ is the target Q-network that is obtained as an exponentially moving average of the Q-network. In discrete action space setting, the value function is given by
\begin{align*}
    &V_{\bar{\psi}}(s_t, G) = \\
    &\sum_{i=1}^{K_t}\pi_w(a_{t,i}|s_t, G) [Q_{\bar{\psi}}(s_t, a_{t,i}, G) - \alpha \log \pi_w(a_{t,i} | s_t, G)]
\end{align*}
where $K_t$ is the number of actions at state $s_t$. Similarly, in discrete setting, the policy network is learned by minimizing the KL divergence between the policy and the exponential of the Q-network
\begin{align*}
    &J_\pi(w) = \mathbb{E}_{(s_t, G) \sim \mathcal{D}} [ \\ &\quad \sum_{i=1}^{K_t}\pi_w(a_{t,i}|s_t, G) [\alpha \log \pi_w(a_{t,i} | s_t, G) - Q_\psi(s_t, a_{t,i}, G)] \\
    &]
\end{align*}
The temperature $\alpha$ is updated by minimizing
\begin{align*}
    J(\alpha) = \sum_{i=1}^{K_t}\pi_w(a_{t,i}|s_t, G) [-\alpha_t \log \pi_w(a_{t,i}|s_t, G) - \alpha_t \bar{\mathcal{H}}]
\end{align*}
where $\bar{\mathcal{H}}$ is the desired minimum expected entropy of $\pi_w(a_t|s_t, G)$.
\end{linenomath}

\noindent \textbf{Discriminator}: Our model is based on the adversarial inverse reinforcement learning (AIRL) framework~\cite{fu2018learning}. The generator in the AIRL framework is the SAC policy model we just described, and the discriminator is defined as follows:
\begin{linenomath}
\postdisplaypenalty=0
\begin{align*}
    &D_{\theta, \phi}(s_t, a_{t, i}, s_{t+1}, G) = \\
    &\quad\quad\quad\quad\quad\quad \frac{\exp \{ f_{\theta, \phi}(s_t, a_{t, i}, s_{t+1}, G) \}}{\exp \{ f_{\theta, \phi}(s_t, a_{t, i}, s_{t+1}, G) \} +\pi(a_{t, i}|s_t, G)}
\end{align*}
where  $f_{\theta, \phi}(s_t, a_{t, i}, s_{t+1}, G) = g_\theta(s_t, a_{t, i}, G) + \gamma h_\phi(s_{t+1}, G) - h_\phi(s_t, G)$.
$g_\theta$ and $h_\phi$ are multilayer feedforward neural network
\begin{align*}
    g(s_t, a_{t, i}, G) &= \text{MLP}([h^s_t; g_t; e^a_{t, i}]) \\
    h(s_t, G) &= \text{MLP}([h^s_t; g_t])
\end{align*}
The parameter $\theta$ and $\phi$ are optimized by minimizing the cross-entropy loss with label smoothing~\cite{szegedy2016rethinking}
\begin{align*}
    L(\theta, \phi) =& \mathbb{E}_{(s_t, a_{t, i}, s_{t+1}, G) \sim \mathcal{D}} [\\
    &\quad\quad\quad(1-\ell)\log D(s_t, a_{t, i}, s_{t+1}, G) \\
    &\quad\quad\quad+\ell \log (1-D(s_t, a_{t, i}, s_{t+1}, G)) \\
    &] \\
    &+\mathbb{E}_{(s_t, a_{t, i}, s_{t+1}, G) \sim \mathcal{D}'} [ \\
    &\quad\quad\quad\ell\log D(s_t, a_{t, i}, s_{t+1}, G) \\
    &\quad\quad\quad+ (1-\ell) \log (1-D(s_t, a_{t, i}, s_{t+1}, G)) \\
    ]
\end{align*}
\end{linenomath}
where $\ell$ is the label smoothing hyper-parameter, and $\mathcal{D}$ and $\mathcal{D}'$ are the positive and negative examples discussed in Section 4.

\noindent \textbf{Variational Goal Generator}: The variational goal generator has a encoder and a decoder. The encoder encodes the sequence of observations in a trajectory $\tau$, and the decoder samples a natural language goal $G$ for that trajectory. Let $e^a_t$ be the embedding of action selected at step $t$. We first use a bi-directional LSTM to encode actions in the trajectory $h^a_1, h^a_2, ..., h^a_T = \text{BiLSTM}(e^a_1, e^a_2, ..., e^a_T)$, then we use $h^a_{t-1}$ to attend over observations at $t$: $\alpha_i^a = \text{Softmax}(\text{Linear}(h_{t-1}^a) \cdot f_{t, i})$. Then the summarized observation feature vector at $t$ is $f_t = \sum_{i=1}^{36} \alpha^a_i f_{t, i}$.
Then the generative process of the $n$-th word in the natural language goal is as follows:
\begin{linenomath}
\postdisplaypenalty=0
\begin{align*}
    h^o_1, h^o_2, ..., h^o_T &= \text{BiLSTM}(f_1, f_2, ..., f_T) \\
    h^o &= \text{ElementWiseMax}(h^o_1, h^o_2, ..., h^o_T) \\
    \mu_{prior}, \sigma^2_{prior} &= \text{MLP}(h^o), \text{Softplus}(\text{MLP}(h^o)) \\
    p(z) &= \mathcal{N}(z|\mu_{prior}, \sigma^2_{prior}\mathbf{I}) \\
    z &\sim \mathcal{N}(\mu_{prior}, \sigma^2_{prior}\mathbf{I}) \\
    h^g_n &= \text{LSTMCell}\left(h^g_{n-1}, e^w_n \right) \\
    \alpha^o_t &= \text{Softmax}\left(\text{Linear}([h^g_n; z]) \cdot h^o_t\right) \\
    p_{n+1}(w) &= \text{Softmax}\left(\text{MLP}\left(\left[\sum_{t=1}^T \alpha^o_t h^o_t; h^g_n; z\right]\right)\right) \\
    w &= \argmax_w p_{n+1}(w)
\end{align*}
where $\text{Softplus}$ is a function $f(x)=\ln(1+e^x)$ to ensure that $\sigma^2$ is positive, $e^w_n$ is the word embedding of $w_n$, and $p_{n+1}(w)$ is the probability of selecting a word $w$ at position $n+1$.
The posterior distribution of the latent variable $z$ is approximated by
\begin{gather*}
    h^g = \text{ElementWiseMax}(h^g_1, h^g_2, ..., h^g_N) \\
    \mu_{posterior}, \sigma^2_{posterior} = \text{MLP}(h^o, h^g), \text{Softplus}(\text{MLP}(h^o, h^g)) \\
    q(z|\tau, G)  = \mathcal{N}(\mu_{posterior}, \sigma^2_{posterior}\mathbf{I})
\end{gather*}
We maximize the variational lower bound $-\lambda D_{KL}(q(z|\tau, G) || p(z)) + \mathbb{E}_{q(z|\tau, G)}[\log p(G|z, \tau)]$. $\lambda$ is KL multiplier that gradually increases from $0$ to $1$ during training. 
\end{linenomath}

\noindent \textbf{Goal-agnostic Policy $\pi_p$}. In Section 3.4, we propose to learn a goal-agnostic policy $\pi_p$ that can sample trajectories in new environments to construct expert demonstrations. We learn this goal-agnostic policy $\pi_p$ using the same network architecture as the base model mentioned above, except that now that the action selection is not conditioned on goals. That is, the probability of selecting an action which originally in Equation \ref{eq:basemodelp} now becomes
\begin{linenomath}
\postdisplaypenalty=0
\begin{align*}
    p^a_i = \text{Softmax}\left(\text{BiLinear}\left(h_t^s, e^a_{t, i}\right)\right)
\end{align*}
\end{linenomath}

\noindent \textbf{Hyper-parameters}: 
By default, the hidden size of MLP in all the models in this paper is $512$, and the number of layers is $2$. The hidden size of LSTM is $512$ ($256$ for BiLSTM) and the number of layer is $1$. The size of word embedding is $300$. The dropout rate is $0.5$ for all the models. The batch sizes for training SAC and the discriminator are both $100$. All the models are optimized by Adam and the learning rate is $1e-4$. The target entropy in SAC is $-\log(1/\text{MAX\_NUM\_ACTIONS}) * 0.1$ where $\text{MAX\_NUM\_ACTIONS}=14$. The size of replay buffer in SAC is $10^6$. The reward discount factor is $0.99$. Our implementation of SAC is based on the rlpyt code base~\cite{stooke2019rlpyt}. We apply label smoothing when optimizing the discriminator. The positive label is $0.95$ and the negative label is $0.05$. The size of the latent variable in variational goal generator is $128$. When training the variational goal generator, we gradually increase the KL multiplier $\lambda$ from $0$ to $1$ during the first $50,000$ gradient updates, and we also apply a word dropout with probability $0.25$. All the experiments are run $8$ times and the average and standard deviation of the success rate are reported. One run includes $1$ million interactions with the environments and takes about $30$ hours to train on a NVIDIA V100 GPU.

\section{Examples of Generated Goals}
\label{appendix:examplegoals}
The following are examples of human-annotated vs.\ GoalGenerator-generated instructions for a trajectory in the Room-2-Room dataset.

Example 1:
\begin{list}{}{\leftmargin=0.75cm}
\item \textbf{Human}: \textit{Exit room through the doorway near the fireplace. Keep right and walk through the hallway, turn right, enter the bedroom and wait near the bed.}
\item \textbf{GoalGenerator}: \textit{Walk through the doorway to the right of the fireplace. Walk past the fireplace and turn right. Walk down the hallway and turn right into the bedroom . Stop in front of the bed.}
\end{list}

Example 2:
\begin{list}{}{\leftmargin=0.75cm}
\item \textbf{Human}: \textit{Turn right and exit the room through the door on the left. Turn left and walk out into the hallway. Turn right and enter bedroom. Walk through the bedroom and into the bathroom. Stop once you are in front of the sink.}
\item \textbf{GoalGenerator}: \textit{Turn around and walk through the doorway. Turn left and walk down the hallway. Turn right at the end of the hall and enter the bathroom. Stop in front of the sink.}
\end{list}

Example 3:
\begin{list}{}{\leftmargin=0.75cm}
\item \textbf{Human}: \textit{Turn around and walk outside through the door behind you. Once outside, turn right and walk to the other end of the pool, At the end of the pool turn right and enter the door back into the house and stop behind the 2 white chairs.}
\item \textbf{GoalGenerator}: \textit{Turn around and walk out of the room. Once out, turn right and walk towards the pool. Once you reach the pool, turn right and walk towards the large glass doors. Once you reach the 2 chairs, turn right and enter the large room. Stop once you reach the couch.}
\end{list}

The following are goals generated by the GoalGenerator for trajectories sampled by the goal-agnostic policy in new environments (Section $3.4$).
\begin{enumerate}
    \item \textit{Walk straight across the room and past the couch. Walk straight until you get to a room with a large green vase. Wait there.}
    \item \textit{Exit the bathroom and turn right. Walk down the hallway and turn right. Walk straight until you get to a kitchen area. Wait near the stove.}
    \item \textit{Walk straight down the hallway and into the room with the large mirror. Walk through the door and stop in front of the table with the plant on it.}
\end{enumerate}

\section{Additional Experiments}
\label{appendix:addexp}
\begin{table*}[]
\centering
\begin{tabular}{cccc|c}
\hline
\begin{tabular}[c]{@{}c@{}}Expert\\ Goal Relabeling\end{tabular} & \begin{tabular}[c]{@{}c@{}}Hindsight\\ Goal Relabeling\end{tabular} & \begin{tabular}[c]{@{}c@{}}Hindsight\\ Goal Sampling\end{tabular} & \begin{tabular}[c]{@{}c@{}}Hindsight\\ Experience Replay\end{tabular} & \begin{tabular}[c]{@{}c@{}}seen validation\\ success rate\end{tabular} \\ \hline
\textendash                                                      & \textendash                                                         & \textendash                                                       & \textendash                                                           & $0.449 \pm 0.0130$                                                     \\ \hline
\ding{51}                                                        & \textendash                                                         & \textendash                                                       & \textendash                                                           & $0.450 \pm 0.0122$                                                     \\
\textendash                                                      & \ding{51}                                                           & \textendash                                                       & \textendash                                                           & $0.427 \pm 0.0048$                                                     \\
\textendash                                                      & \textendash                                                         & \ding{51}                                                         & \textendash                                                           & $0.399 \pm 0.0124$                                                     \\
\textendash                                                      & \textendash                                                         & \textendash                                                       & \ding{51}                                                             & $0.464 \pm 0.0126$                                                     \\ \hline
\ding{51}                                                        & \ding{51}                                                           & \textendash                                                       & \textendash                                                           & $0.504 \pm 0.0185$                                                     \\
\ding{51}                                                        & \textendash                                                         & \ding{51}                                                         & \textendash                                                           & $0.444 \pm 0.0134$                                                     \\
\ding{51}                                                        & \textendash                                                         & \textendash                                                       & \ding{51}                                                             & $0.459 \pm 0.0050$                                                    \\
\textendash                                                      & \ding{51}                                                           & \ding{51}                                                         & \textendash                                                           & $0.503 \pm 0.0117$                                                     \\
\textendash                                                      & \ding{51}                                                           & \textendash                                                       & \ding{51}                                                             & $0.101 \pm 0.0790$                                                     \\
\textendash                                                      & \textendash                                                         & \ding{51}                                                         & \ding{51}                                                             & $0.378 \pm 0.0192$                                                     \\ \hline
\ding{51}                                                        & \ding{51}                                                           & \ding{51}                                                         & \textendash                                                           & $\mathbf{0.530 \pm 0.0138}$                                            \\
\ding{51}                                                        & \ding{51}                                                           & \textendash                                                       & \ding{51}                                                             & $0.491  \pm 0.0146$                                                    \\
\ding{51}                                                        & \textendash                                                         & \ding{51}                                                         & \ding{51}                                                             & $0.452 \pm 0.0088$                                                     \\
\textendash                                                      & \ding{51}                                                           & \ding{51}                                                         & \ding{51}                                                             & $0.476 \pm 0.0166$                                                     \\ \hline
\ding{51}                                                        & \ding{51}                                                           & \ding{51}                                                         & \ding{51}                                                             & $0.521 \pm 0.0093$                                                    \\ \hline
\end{tabular}
\caption{A complete ablation study of EGR, HGR, HGS and HER on seen validation set.}
\label{tb:fullablation}
\end{table*}
\begin{table*}[]
\centering
\begin{tabular}{l|cccc}
\hline
               & 5,000               & 10,000             & 15,000             & 20,000             \\ \hline
RandomExplore & $0.274 \pm 0.0086$ & $0.303 \pm 0.0092$ & $0.347 \pm 0.0098$ & $0.348 \pm 0.0115$ \\
LangGoalIRL\_SS\_G2 & $0.430 \pm 0.0080$ & $0.459 \pm 0.0084$ & $0.483 \pm 0.0124$ & $0.490 \pm 0.0068$ \\
LangGoalIRL\_SS    & $0.452 \pm 0.0067$ & $0.475 \pm 0.0062$ & $0.491 \pm 0.0065$ & $\mathbf{0.496 \pm 0.0039}$ \\ \hline
\end{tabular}
\caption{The success rate in unseen validation with different self-supervised learning setting. Algorithms sample $5,000$ to $20,000$ demonstrations in unseen environments. RandomExplore samples demonstrations by randomly select an unvisited nearby viewpoint, while LangGoalIRL\_SS and LangGoalIRL\_SS\_G2 sample demonstrations using goal-agnostic policy $\pi_p$. During fine-tuning, LangGoalIRL\_SS\_G2 samples goals only from $\{(\bm{E}, \bm{G}^v) | v \le 2\}$, while the other two algorithms sample goals from $\{(\bm{E}, \bm{G}^v) | v \ge 1\}$.}
\label{tb:ss3}
\end{table*}
\begin{table*}[]
\centering
\begin{tabular}{l|ccccc}
\hline
                 & 0                  & 5000               & 10,000             & 15,000              & 20,000             \\ \hline
LangGoalIRL\_BASE & $0.449 \pm 0.0130$ & $0.469 \pm 0.0111$ & $0.472 \pm 0.0105$ & $0.478 \pm 0.0108$  & $0.481 \pm 0.0156$ \\
LangGoalIRL         & $0.530 \pm 0.0138$ & $0.532 \pm 0.0126$ & $0.535 \pm 0.0113$ & $0.538 \pm  0.0155$ & $\mathbf{0.543 \pm 0.0125}$ \\ \hline
\end{tabular}
\caption{Self-supervised learning can also be applied to seen validation set. This table shows the success rate in seen validation when self-supervised learning is used to augment the expert demonstrations in seen validation. We use the goal-agnostic policy $\pi_p$ to sample $5,000$ to $20,000$ trajectories, relabel them with goal generator, and add them to the seen validation for training.}
\label{tb:sstrain}
\end{table*}
\begin{table*}[]
\centering
\begin{tabular}{l|cccc}
\hline
\#relabeled expert goals. & 0 & 1                  & 2                   & 3                   \\ \hline
LangGoalIRL           &    $0.503\pm0.0117$    & $0.525 \pm 0.0090$ & $\mathbf{0.530 \pm  0.0138}$ & $0.528 \pm 0.0105$  \\
LangGoal\_ONLY\_EGR    &   $0.449 \pm 0.0130$      & $0.446 \pm 0.0169$ & $0.450 \pm 0.0122$  & $0.450 \pm  0.0155$ \\ \hline
\end{tabular}
\caption{Success rate in seen validation given different number of relabeled expert goals. This corresponds to setting $N$ in Section 3.3.1 to $0$, $1$, $2$ and $3$.}
\label{tb:expg}
\end{table*}

\begin{table*}[]
\centering
\begin{tabular}{l|cccc}
\hline
                  & BLEU  & Unigram Precision & Bigram Precision & Trigram Precision \\ \hline
seen validation   & 0.341 & 0.804             & 0.478            & 0.259             \\ \hline
unseen validation & 0.316 & 0.791             & 0.453            & 0.234             \\ \hline
\end{tabular}
\caption{BLEU score and n-gram precision of the variational goal generator evaluated on both seen and unseen validation sets.}
\label{tb:bleu}
\end{table*}
Table \ref{tb:fullablation} shows the full ablation study results which also include hindsight experience replay. Notably, when apply HER together with HGR, the success rate drops to $0.101$. The reason is that without EGR and HGS, the discriminator over-estimate the rewards under relabeled goals (as relabeled goals only appear in positive examples of the discriminator), and hence cannot give a good estimate of rewards to trajectories relabeled by HER.

Table \ref{tb:ss3} shows the policy performance in unseen validation with three different self-supervised learning settings. RandomExplore randomly select actions instead of using $\pi_p$ when generating demonstrations in unseen environments (but avoid visiting a viewpoint twice). After collecting demonstrations, RandomExplore fine-tunes the policy in the same way as LangGoalIRL\_SS. LangGoalIRL\_SS\_G2 is different from LangGoalIRL\_SS in that it does not iteratively sample goals from $\{(\bm{E}, \bm{G}^v) | v > 2\}$ as described in Section 3.3.3, so the policy experience a less diverse set of natural language goals compared with LangGoalIRL\_SS. The number of generated demonstrations in unseen environments is set to $5,000$, $10,000$, $15,000$ or $20,000$.
We can see that LangGoalIRL\_SS performs much better than RandomExplore, showing the importance of collecting demonstrations by $\pi_p$ instead of a random policy. LangGoalIRL\_SS also performs better than LangGoalIRL\_SS\_G2, however, the gap between LangGoalIRL\_SS and LangGoalIRL\_SS\_G2 decreases as the number of generated demonstrations increases.

Table \ref{tb:sstrain} shows the success rate of LangGoalIRL and LangGoalIRL\_BASE when self-supervised learning is applied to existing environments. More specifically, in Section 3.4 we propose a self-supervised learning algorithm in new environments (unseen validation set), we can also apply the same idea to existing environments (seen validation set). To do so, we use the goal-agnostic policy $\pi_p$ to sample trajectories in existing environments and relabel them with the variational goal generator. Then we augment the seen validation set with these generated demonstrations. During the training, EGR is only applied to the original expert demonstrations in the seen validation set. From the results we can see that self-supervised learning improves the performance of LangGoalIRL and LangGoalIRL\_BASE in seen validation. Self-supervised learning has a larger impact on LangGoalIRL\_BASE than on LangGaolIRL. This is reasonable since LangGoalIRL can already generalize the policy and reward function by our proposed three strategies (EGR, HGR and HGS).

Table \ref{tb:expg} shows the success rate of LangGoalIRL in seen validation when we vary the number of relabeled expert goals ($N$ in Section 3.3.1). We can see that the policy performances are comparable when the number of relabeled expert goals $N\ge1$.

Table \ref{tb:bleu} shows the BLEU score and n-gram precision of the varitaional goal generator on both seen and unseen environments.

\end{document}